\documentclass[12pt]{colt2022} %
\usepackage{algorithm}
\usepackage{cleveref}

\title[Hardness of Online Private Learning]{Do you pay for Privacy in Online learning?}
\usepackage{times}
\usepackage{amartya_ltx}
\usepackage{natbib}
\usepackage{tikz}%

\coltauthor{%
 \Name{Giorgia Ramponi} \Email{giorgia.ramponi@ai.ethz.ch}\\
 \addr ETH AI Center, Zurich, Switzerland
 \AND
 \Name{Amartya Sanyal} \Email{amartya.sanyal@ai.ethz.ch}\\
 \addr ETH AI Center, Zurich, Switzerland
}
\definecolor{mygreen}{rgb}{0.0, 0.5, 0.0}

\begin{document}

\maketitle

Online learning, in the mistake bound model, is one of the most
fundamental concepts in learning theory. Differential privacy,
instead, is the most widely used statistical concept of privacy in the
machine learning community. It is thus clear that defining learning
problems that are online differentially privately learnable is of
great interest.  In this paper, we pose the question on if the two
problems are equivalent from a learning perspective, i.e., is privacy
for free in the online learning framework? 

\section{Introduction}
\label{sec:intro}

{\em Online learning}, in the mistake bound model, is one of the most
fundamental concepts in learning theory. Let $X=\bigcup X_n$ be the
instance space. The learner, \(\cA\), in this model, receives at each
timestep \(t\) an unlabelled example \(x_t\in X\), predicts a label
\(\widehat{y}_t\) corresponding to \(x_t\), and then receives the true
label \(y_t\) for \(x_t\). During this interaction, the learner
maintains a working hypothesis \(h_t =
\cA\br{\bc{\br{x_{\tau},y_{\tau}}}_{\tau=1}^{t-1}}\), which it uses to
predict \(\widehat{y}_t=h_t(x_t)\), and then uses the true label
\(y_t\) to update the working hypothesis to \(h_{t+1}\). The
performance of the learner \(\cA\) is measured by the number of
\textit{mistakes} it makes, i.e.,:\looseness=-1
\begin{align}\label{eq:mistakes}
    \text{Mistakes}\left(\cA,T, \left(x_t, y_t\right)_{t=1}^\infty\right) := \sum_{t=1}^T \left(h_t(x_t) \ne y_t\right).
\end{align}
Given this definition of performance, a hypothesis class \(\cC\) on the instance space \(X=\bigcup X_n\) is said to be {\em online learnable} in the mistake bound model if there exists a learner \(L\) that makes at most \(\mathrm{poly}\br{n,size\br{c}}\) mistakes on any sequence of samples consistent with a concept \(c\in\cC\), where \(p\) is some polynomial. This is also known as the {\em realisable setting}.

Another relevant concept in learning theory is privacy. The most widely used statistical notion of privacy in the machine learning literature is \textit{differential privacy}. An \(\br{\epsilon,\delta}\)-differentially private~(randomised) algorithm is guaranteed to output {\em similar} distributions over the output space of the algorithm when presented with inputs that only differ in one element. More formally, in the offline setting, a learning algorithm \(\cA:\cX\rightarrow\cY\) is said to be \(\br{\epsilon,\delta}\)-differentially private if, for any two datasets \(S_1,S_2\) that differ in just one element, we have that \(\bP\bs{\cA\br{S_1}\in Q}\leq e^\epsilon\bP\bs{\cA\br{S_2}\in Q} + \delta\) where \(Q\subseteq \cY\) is any subset of the output space of the algorithm. We define differential privacy in the online setting in~\Cref{defn:online-dp}.

Some previous works \citep{jain2012differentially,agarwal2017price,abernethy2019online} treat the problem of constructing online learning algorithms (mostly in the regret minimization setting \citep{shalev2007online}) maintaining the differential-privacy properties. However, it is still not clear how these two problems (non-private mistake bound and private mistake-bound) are connected and if there exists some problem which is online learnable in the mistake bound model but not private online learnable in the mistake bound model. In other words, the open problem presented in this paper concerns a fundamental question about learning: 
\begin{center}
    ``Is privacy for free in the online learning framework?''
\end{center}

\section{Related works on learnability}
\tikzset{every picture/.style={line width=0.75pt}} %

\tikzset{every picture/.style={line width=0.75pt}} %
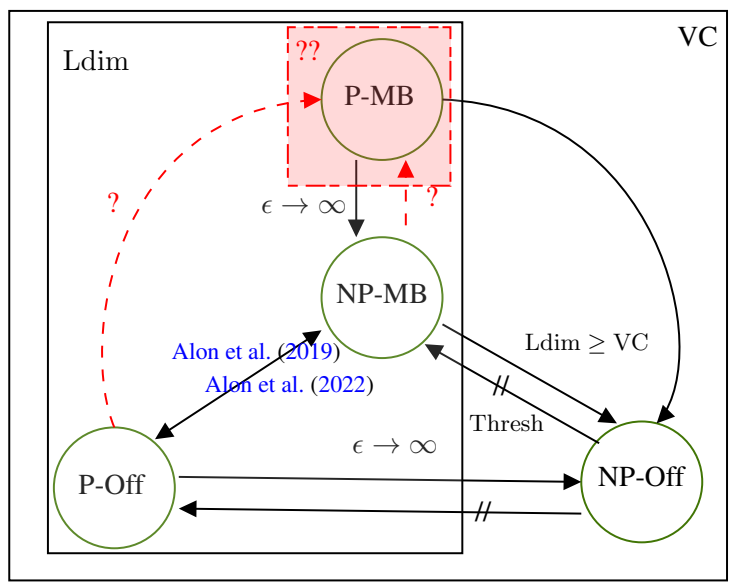
\begin{figure}\centering
\begin{tikzpicture}[x=0.75pt,y=0.75pt,yscale=-1,xscale=1]

\draw  [fill={rgb, 255:red, 155; green, 155; blue, 155 }  ,fill opacity=0. ] (19.5,86.77) -- (381.5,86.77) -- (381.5,374.23) -- (19.5,374.23) -- cycle ;
\draw  [color={rgb, 255:red, 65; green, 117; blue, 5 }  ,draw opacity=1 ] (177,231.5) .. controls (177,214.66) and (190.66,201) .. (207.5,201) .. controls (224.34,201) and (238,214.66) .. (238,231.5) .. controls (238,248.34) and (224.34,262) .. (207.5,262) .. controls (190.66,262) and (177,248.34) .. (177,231.5) -- cycle ;
\draw  [color={rgb, 255:red, 65; green, 117; blue, 5 }  ,draw opacity=1 ] (177,131.5) .. controls (177,114.66) and (190.66,101) .. (207.5,101) .. controls (224.34,101) and (238,114.66) .. (238,131.5) .. controls (238,148.34) and (224.34,162) .. (207.5,162) .. controls (190.66,162) and (177,148.34) .. (177,131.5) -- cycle ;
\draw  [color={rgb, 255:red, 65; green, 117; blue, 5 }  ,draw opacity=1 ] (308,324.5) .. controls (308,307.66) and (321.66,294) .. (338.5,294) .. controls (355.34,294) and (369,307.66) .. (369,324.5) .. controls (369,341.34) and (355.34,355) .. (338.5,355) .. controls (321.66,355) and (308,341.34) .. (308,324.5) -- cycle ;
\draw  [color={rgb, 255:red, 65; green, 117; blue, 5 }  ,draw opacity=1 ] (42,327.5) .. controls (42,310.66) and (55.66,297) .. (72.5,297) .. controls (89.34,297) and (103,310.66) .. (103,327.5) .. controls (103,344.34) and (89.34,358) .. (72.5,358) .. controls (55.66,358) and (42,344.34) .. (42,327.5) -- cycle ;

\draw (187,123) node [anchor=north west][inner sep=0.75pt]   [align=left] {P-MB};
\draw (183,223) node [anchor=north west][inner sep=0.75pt]   [align=left] {NP-MB};
\draw (315,315) node [anchor=north west][inner sep=0.75pt]   [align=left] {NP-Off};
\draw (53,319) node [anchor=north west][inner sep=0.75pt]
[align=left] {P-Off};

\draw (191,302) node [anchor=north west][inner sep=0.75pt]   [align=left] {$\epsilon \rightarrow \infty$};

\draw    (105,321.97) -- (305,324.46) ;
\draw [shift={(308,324.5)}, rotate = 180.71] [fill={rgb, 255:red, 0;
green, 0; blue, 0 }  ][line width=0.08]  [draw opacity=0] (8.93,-4.29)
-- (0,0) -- (8.93,4.29) -- cycle    ;

\draw (145,181) node [anchor=north west][inner sep=0.75pt]
[align=left] {$\epsilon \rightarrow \infty$};

\draw    (194.5,162) -- (194.5,198) ;
\draw [shift={(194.5,201)}, rotate = 270] [fill={rgb, 255:red, 0;
green, 0; blue, 0 }  ][line width=0.08]  [draw opacity=0] (8.93,-4.29)
-- (0,0) -- (8.93,4.29) -- cycle    ;

\draw    (238,244.97) -- (323.39,293.49) ;
\draw [shift={(326,294.97)}, rotate = 209.6] [fill={rgb, 255:red, 0;
green, 0; blue, 0 }  ][line width=0.08]  [draw opacity=0] (8.93,-4.29)
-- (0,0) -- (8.93,4.29) -- cycle    ;

\draw (278,248) node [anchor=north west][inner sep=0.75pt]
[align=left] {\footnotesize\(\mathrm{Ldim}\geq \mathrm{VC}\)};

\draw    (348.14,292.47) .. controls (374.65,257.65) and
(347.2,135.86) .. (238,131.5) ;

\draw [shift={(346,294.97)}, rotate = 314.06] [fill={rgb, 255:red, 0;
green, 0; blue, 0 }  ][line width=0.08]  [draw opacity=0] (8.93,-4.29)
-- (0,0) -- (8.93,4.29) -- cycle    ;

\draw    (231.61,256.45) -- (317,304.97) ; 
\draw [shift={(229,254.97)}, rotate = 29.6] [fill={rgb, 255:red, 0;
green, 0; blue, 0 }  ][line width=0.08]  [draw opacity=0] (8.93,-4.29)
-- (0,0) -- (8.93,4.29) -- cycle    ;
\draw (250,289) node [anchor=north west][inner sep=0.75pt]
[align=left] {\footnotesize\(\mathrm{Thresh}\)};

\draw (262,269) node [anchor=north west][inner sep=0.75pt]   [align=left] {//};

\draw  [fill={rgb, 255:red, 255; green, 255; blue, 255 }  ,fill
opacity=0.15 ]%
(39,92.5) -- (248,92.5) -- (248,360.5) -- (39,360.5) -- cycle ;

\draw (355,93) node [anchor=north west][inner sep=0.75pt]
[align=left] {VC};

\draw    (174.49,249.61) -- (95.51,301.33) ;

\draw [shift={(177,247.97)}, rotate = 146.78] [fill={rgb, 255:red, 0;
green, 0; blue, 0 }  ][line width=0.08]  [draw opacity=0] (8.93,-4.29)
-- (0,0) -- (8.93,4.29) -- cycle    ;

\draw (100,253) node [anchor=north west][inner sep=0.75pt]   [align=right] {\footnotesize\cite{alon2019private}};

\draw (45,105) node [anchor=north west][inner sep=0.75pt]
[align=left] {\(\mathrm{Ldim}\)};

\draw (253,333) node [anchor=north west][inner sep=0.75pt]
[align=left] {//};

\draw    (108,338.01) -- (308,340.5) ;
\draw [shift={(105,337.97)}, rotate = 0.71] [fill={rgb, 255:red, 0; green, 0; blue, 0 }  ][line width=0.08]  [draw opacity=0] (8.93,-4.29) -- (0,0) -- (8.93,4.29) -- cycle    ;

\draw [shift={(93,302.97)}, rotate = 326.78] [fill={rgb, 255:red, 0;
green, 0; blue, 0 }  ][line width=0.08]  [draw opacity=0]
(8.93,-4.29) -- (0,0) -- (8.93,4.29) -- cycle    ;

\draw (205,283) node [anchor=south east][inner sep=0.75pt]   [align=left] {\footnotesize\cite{alon2022private}};

\draw  [color=red, dash pattern={on 4.5pt off 4.5pt}]  (72.5,297) .. controls (55.09,256.17) and (69.85,137.17) .. (175.4,131.58) ;
\draw [shift={(177,131.5)}, rotate = 177.61] [fill={rgb, 255:red, 255; green, 0; blue, 0 }  ][line width=0.08]  [draw opacity=0] (8.93,-4.29) -- (0,0) -- (8.93,4.29) -- cycle    ;

\draw  [color=red, dash pattern={on 4.5pt off 4.5pt}]  (219.5,165) -- (219.5,201) ;
\draw [shift={(219.5,162)}, rotate = 90] [fill={rgb, 255:red, 255; green, 0; blue, 0 }  ][line width=0.08]  [draw opacity=0] (8.93,-4.29) -- (0,0) -- (8.93,4.29) -- cycle    ;

\draw (228,174) node [anchor=north west][inner sep=0.75pt]   [align=left] {\color{red}?};
\draw (67,177) node [anchor=north west][inner sep=0.75pt]   [align=left] {\color{red}?};

\draw  [fill={rgb, 255:red, 255; green, 255; blue, 255 }  ,fill
opacity=0.15 ][color=red, dash pattern={on 3.75pt off 3pt on 7.5pt off 1.5pt}]
(160,95) -- (242,95) -- (242,175) -- (160,175) -- cycle ;

\draw (162,100) node [anchor=north west][inner sep=0.75pt]
[align=left] {\color{red}??};

\end{tikzpicture}
\caption{The figure summerizes the relation between the mentioned four online learning problems: Non-Private Oflline Learning (NP-Off), Non-Private Mistake Bound Learning (NP-MB), Private-Offline Learning (P-Off) and Private Mistake-Bound model (P-MB).}
\label{fig:learningproblems}
\end{figure}

In this section we discuss existing literature on characterising the
learning problems introduced above and establish the connections
between them. We summarise these relations
in~\Cref{fig:learningproblems}.\looseness=-1

\paragraph{Non-Private Offline Learning} The (non-private) offline learning (NP-off) is the most classical learning problem in learning theory. This was formalised by the seminal paper of~\citet{valiant1984theory} as {\em Probably approximately correct}~(PAC) learnability. A hypothesis class is said to be \(\br{\alpha,\beta}\)-PAC learnable if there is an algorithm that when given access to a number of samples polynomial in \(\frac{1}{\alpha},\frac{1}{\beta}\), and the problem size returns a hypothesis that achieves error less than \(\alpha\) with probability greater than \(1-\beta\). Here, the problem size simply refers to the minimal size of a representation of a hypothesis from the hypothesis class. It is now well known that the Vapnik–Chervonenkis dimension~(VC) dimension~\citep{vapnik1999nature} exactly characterises non-private offline learnability in that any hypothesis class with finite VC dimension is learnable in the PAC model (and vice versa).

\paragraph{Non-Private Online Learning}
As discussed before, a hypothesis class \(\cH\) is said to be learnable in the online mistake bound model if there is a finite \(M\) and an online algorithm \(\cA\) such that \(\cA\) makes at most \(M\) mistakes on any sequence of data labelled with some \(h\in\cH\). Interestingly, it is also possible to characterise online learnability using a different combinatorial measure of the hypothesis class called the Littlestone dimension, which we define in~\Cref{defn:ldim}.~\citet{littlestone1988learning} proved that for any hypothesis class \(\cH\), there exists an online learning algorithm that makes at most \(\mathrm{Ldim}\br{\cH}\) mistakes on any sequence labelled by some \(h\in\cH\), thereby characterising online learnability.\looseness=-1

\begin{defn}[Littlestone dimension~\citep{littlestone1988learning}]~\label{defn:ldim}
The littlestone dimension of a hypothesis class \(\cH\), denoted as \(\mathrm{Ldim}\br{\cH}\) is the depth of the largest tree that can be shattered by \(\cH\), where we define {\em ``shattering a tree''} in~\Cref{defn:shattering}.
\end{defn}

\begin{defn}[Shattering a tree]\label{defn:shattering}
Consider a full binary tree of depth \(d\) such that each node is labelled by some \(x\in\cX\). For a set of labels \(\bc{y_i}_{i=1}^{d}\), define its corresponding path as starting from the root and taking the left child when \(y=-1\) and the right child when \(y=+1\). The tree is said to be shattered by some \(h\in\cH\) if for every set of labels in \(\bc{-1,1}^d\), its corresponding path can be shattered by some \(h\in\cH\) i.e. for all \(x_i\) in the path, \(h\br{x_i}=y_1\). 
\end{defn}

\paragraph{Private Offline Learning}
The non-private offline PAC learnability problem was extended to the
case of differentially private learnability
by~\citet{raskhodnikova2008can}. A hypothesis  class is
\(\br{\epsilon,\delta,\alpha,\beta}\)-differentially private PAC
learnable if there exists an \(\br{\alpha,\beta}\)-PAC learning
algorithm that is also \(\br{\epsilon,\delta}\)-differentially
private.~\citet{raskhodnikova2008can} showed that any problem that is
PAC learnable is also learnable by a differentially private learning
algorithm but the required number of samplesdepends on the size
of the input space in addition to the VC dimension, which can be
arbitrarily larger than the VC dimension. This left open the question
of whether the sample complexity can be characterised exactly by a
combinatorial measure of the complexity of the hypothesis class.

~\citet{alon2019private} resolved the question partially by proving
that the required number of samples is at least
\(\Omega\br{\log^*\br{\mathrm{Ldim\br{\cH}}}}\) where \(\log^{*}\) is
the iterated logarithm.~\citet{alon2022private} showed the reverse
side and concluded that any class with finite littlestone dimension
can be learned offline privately  with a finite number of samples.
Specifically they showed that any hypothesis class with a finite
Littlestone dimension \(d\) is private learnable with number of
samples doubly exponential in \(d\). This concludes that private
offline learnability is exactly characterised by the littlestone
dimension, which in turn exactly characterises online learnability in
the mistake bound model thereby showing an equivalence between the two
regimes. However, the question remains open whether private online
learnability, with a suitable definition, is harder than non-private
online learnability.\looseness=-1

\section{Open Problem}
In this section we expose the research question introduced in this paper. Before it, we introduce the concept of $\{\epsilon,\delta\}$-differentially private online learning algorithm.
\begin{defn}[$\{\epsilon,\delta\}$-differential online privacy]\label{defn:online-dp}
Let $\mathcal{H}$ be a set of hypotheses \(\cH=\bigcup_{n=1}^{\infty}\cH_n\) over the input space \(\cX=\bigcup_{n=1}^{\infty} \cX_n\). Then an online algorithm $\mathcal{A}$ is $\{\epsilon,\delta\}$-online differentially private if  for all \(T\in\bN\), for any two sequences of points \(S_T\) and \(S'_T\) that differs in at most one entry the following holds:
\begin{equation*}
    \text{Pr}(\mathcal{A}(S_T) \in \mathcal{S}) \le e^\epsilon\text{Pr}(\mathcal{A}(S'_T) \in \mathcal{S}) + \delta
\end{equation*}
\end{defn}
The question that we pose is if every problem that is online learnable it is also online privately learnable, in other words, if the set of problems solvable in these two learnability classes are the same. This question can be solved proving one of the two following theorems, where theorem \ref{thm:imposs} implies that there exists a problem which is online learnable but non-online private learnable and theorem \ref{thm:poss} implies, instead, the opposite.
\begin{thm}\label{thm:imposs}
There exists a set of hypotheses \(\cH=\bigcup_{n=1}^{\infty}\cH_n\) over the input space \(\cX=\bigcup_{n=1}^{\infty} \cX_n\) such that for all \(T\in\bN\), for any sequence of points \(S_T=\bc{\br{x_1, h^*\br{x_1}},\ldots, \br{x_T,h^*\br{x_T}}}\), such that \(h^*\in\cH\),\begin{enumerate}
    \item~(Online learnable) there exists an online algorithm \(\cA\)
    that does not make more than \(M\) mistakes~(\cref{eq:mistakes})
    on the sequence \(S_T\) for some \(M<\infty\).
    \item~(Not privately online learnable) any \(\br{\epsilon,\delta}\)-differentially private online algorithm makes at least \(M'\ge M + \alpha\br{\epsilon,\delta,T}\) mistakes,
\end{enumerate}  
where \(\alpha:\reals\times\bs{0,1}\rightarrow\bN\) is such that $\alpha\br{\epsilon,\delta,T} \gtrsim_\delta \frac{\sqrt{T}}{\epsilon}
$.
\end{thm}
~\Cref{thm:imposs} claims that there exists some hypothesis class that
is non-privately online learnable but any private online algorithm
makes infinite mistakes when \(\epsilon\lesssim\sqrt{T}\). Here, the
symbols \(\gtrsim\) and \(\lesssim\) mean greater than or less than up
to a multiplicative constant and \(\gtrsim_{\delta}\) ignores the
dependance on \(\delta\). As we know that a non-private algorithm can
solve the problem with small number of mistakes, it is natural to
expect that any hardness result would only hold for a sufficiently
small \(\epsilon\). We next state another hypothesis  which states
that any non-privately online learnable hypothesis class is also
privately online learnable. We state this in~\Cref{thm:poss}
below.\looseness=-1

\begin{thm}\label{thm:poss} Let \(\cH\) be any online learnable
hypothesis class. Then, for all \(h^*\in\cH,T\in\bN,
\epsilon,\delta>0\),  there exists an
\(\br{\epsilon,\delta}\)-differentially private online algorithm
\(\cA\) that makes a finite number of mistakes, as defined
in~\Cref{eq:mistakes},~for any sequence of points \(S_T=\bc{\br{x_1,
h^*\br{x_1}},\ldots, \br{x_T,h^*\br{x_T}}}\) of length \(T\) labelled
by \(h^*\in\cH\).
\end{thm}
Theorem \ref{thm:poss} states, instead, that for every hypothesis
class that is online learnable, there also exists an $(\epsilon,
\delta)$-differentially private online learning algorithm by which it
is online learnable.

By definition, all privately online learnable problems are also
non-privately online learnable~(\(\epsilon\to 0\)). Therefore, one
possible implication of a proof for~\Cref{thm:imposs} would be the
definition of a combinatorial measure that is even more restrictive
than littlestone definition, which we are not aware of and is perhaps
of even wider interest to the learning theory community. However,
recent results from~\citet{bousquet2021theory} in the context of {\em
universal learning}~(which is another definition of learnability in
the same spirit as PAC learning) suggests that a combinatorial measure
that is more restrictive than the Littlestone dimension is unlikely.
In particular, they show that there are only three possible rates in
universal learning with the fastest being characterised by the
littlestone dimension and the slowest by VC dimension. This
makes~\Cref{thm:poss} more likely. \looseness=-1

Some initial progress towards this has been made
by~\citet{golowich2021littlestone} who proved, in the oblivious
setting, that the number of mistakes grows logarithmically in \(T\).
This, in fact, disproves~\cref{thm:poss} for the setting of oblivious
adversaries~\footnote{In~\citet{golowich2021littlestone} the authors
also provide some results also for the setting of adaptive
adversaries, but under a definition of differential privacy more
suited to the setting of adaptive adversaries}. Further, even for
oblivious adversaries, it is possible to ask whether~\Cref{thm:poss}
can be proved for \(\alpha\) being any monotonically increasing
function in \(T\). An interesting outcome of a proof
for~\Cref{thm:poss} is a general algorithm to convert an online
learner to a private online learner. We promise a wheel of parmigiano
reggiano to whoever proves~\Cref{thm:imposs} or a tub of biriyani for
solving~\Cref{thm:poss}.
\bibliography{references}
\end{document}